\documentclass[journal]{IEEEtran}
\usepackage[a4paper, left=1in, right=1in, top=1in, bottom=1in]{geometry}
\usepackage{amssymb}
\usepackage{amsmath}
\usepackage{booktabs}
\usepackage{geometry}
\usepackage[font=small]{caption}
\usepackage{subcaption}
\usepackage{graphicx}  
\usepackage{float}     
\usepackage{cite}
\usepackage{amsthm}
\usepackage[hidelinks]{hyperref}

\usepackage{xurl}

\title{\LARGE \bf
Data-driven tool wear prediction in milling, based on a process-integrated single-sensor approach
}
\author{Eric Hirsch and Christian Friedrich$^{1}$
\thanks{$^{1}$ All authors are with the Institute of Applied Research,
Karlsruhe University of Applied Sciences (HKA), Germany,
{\tt\small \{eric.hirsch, christian.friedrich\}@h-ka.de}}%
}

\begin{document}

\maketitle
\thispagestyle{plain}
\pagestyle{plain}

\begin{abstract}

Accurate tool wear prediction is essential for maintaining productivity and minimizing costs in machining. However, the complex nature of the tool wear process poses significant challenges to achieving reliable predictions. This study explores data-driven methods, in particular deep learning, for tool wear prediction. Traditional data-driven approaches often focus on a single process, relying on multi-sensor setups and extensive data generation, which limits generalization to new settings. Moreover, multi-sensor integration is often impractical in industrial environments. To address these limitations, this research investigates the transferability of predictive models using minimal training data, validated across two processes. Furthermore, it uses a simple setup with a single acceleration sensor to establish a low-cost data generation approach that facilitates the generalization of models to other processes via transfer learning. The study evaluates several machine learning models, including transformer-inspired convolutional neural networks (CNN), long short-term memory networks (LSTM), support vector machines (SVM), and decision trees, trained on different input formats such as feature vectors and short-time Fourier transform (STFT). The performance of the models is evaluated on two machines and on different amounts of training data, including scenarios with significantly reduced datasets, providing insight into their effectiveness under constrained data conditions. The results demonstrate the potential of specific models and configurations for effective tool wear prediction, contributing to the development of more adaptable and efficient predictive maintenance strategies in machining. Notably, the ConvNeXt model has an exceptional performance, achieving 99.1\% accuracy in identifying tool wear using data from only four milling tools operated until they are worn.

\end{abstract}






\section{Introduction}
\label{sec1}
\begin{figure*}[htbp]
    \centering
    \includegraphics[width = 0.97 \textwidth]{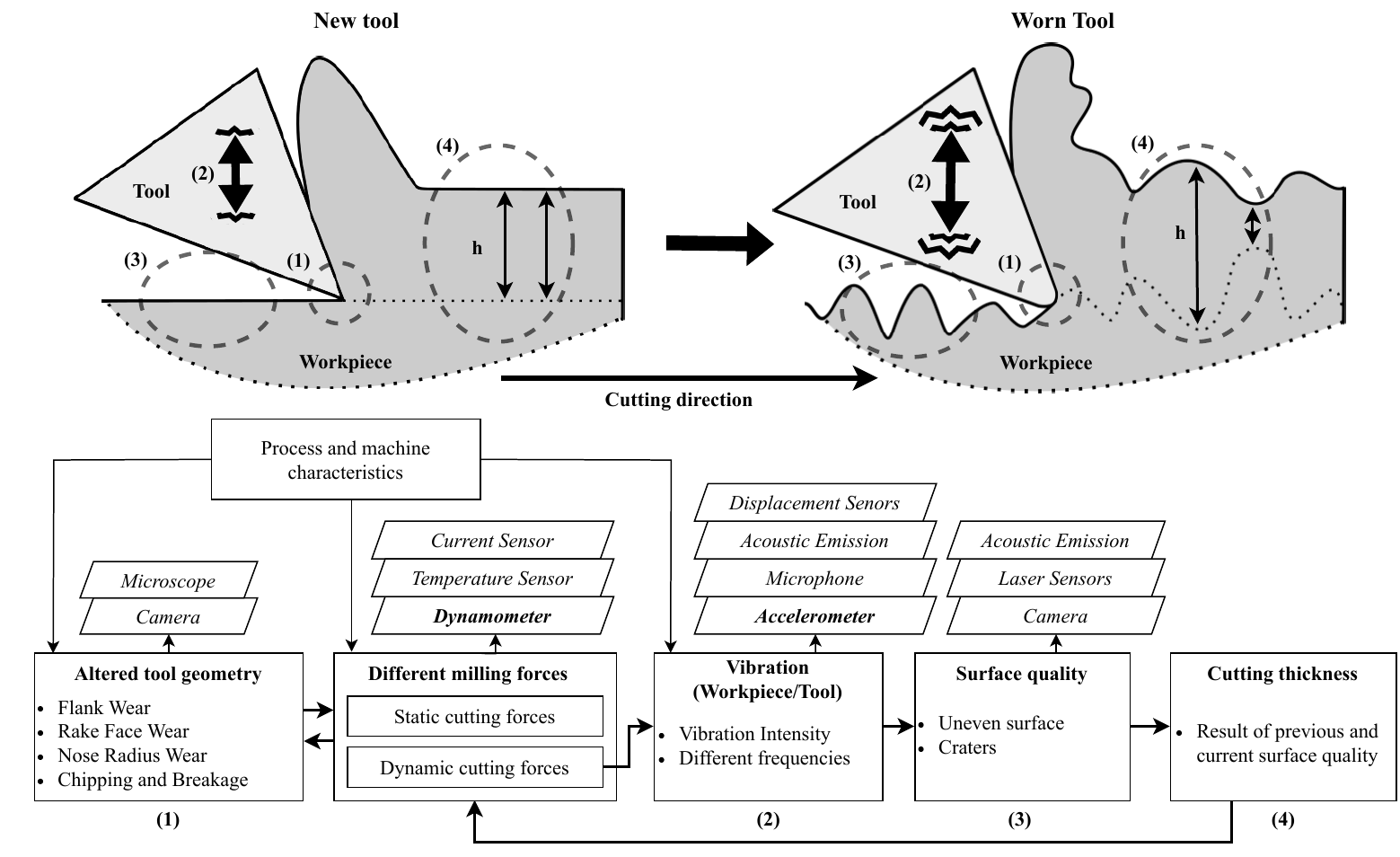}
    \caption{The influence of tool wear and the sensors that can be employed for its monitoring.}
    \label{fig:Sensors_and_wear}
\end{figure*}

 Economically, optimizing tool change times and extending tool life are essential to minimize downtime and maximize output, reducing overall operating costs. A primary concern in this domain is tool wear, which represents a challenge within the machining industry. At present, tool wear is primarily based on supplier recommendations and manual expertise \cite{limDesignInformaticsbasedServices2018}. However, because numerous factors influence tool wear during machining, its behavior is highly complex and variable, making it difficult to predict accurately using conventional or heuristic methods. So, tool wear still occurs, often leading to disruptions in the manufacturing process. Ineffective management of tool wear, whether through premature or delayed identification, can result in a number of unwanted consequences, including unnecessary replacement costs and poor product quality \cite{zengNotchWearPrediction2018}.
Therefore, comprehensive research and development of intelligent tool condition monitoring (TCM) systems \cite{zhangReviewPhysicsBasedDataDriven2024, pimenovArtificialIntelligenceSystems2022,serinReviewToolCondition2020} is essential. TCM systems estimate the tool condition by passing sensor inputs through defined models, assisting machine operators in optimizing their work practices. The tool condition is characterized either through categorical classification, indicating distinct wear states, or through regression to estimate the wear condition. For regression, flank wear is usually applied as a measured variable, which is attempted to be predicted. TCM can operate offline or online \cite{daicuMethodologyMeasuringCutting2022}:
\begin{itemize}
    \item Offline TCM, predictions are made post-process using additional data sources, which may include direct measurement methods. Tool wear is only identified with a significant delay or the process must be temporarily interrupted.
    \item Online TCM offers the process to be continuously evaluated without interrupting the process. This enables short-interval or real-time updates on tool wear. Indirect measurement methods are typically the only viable option in online TCM.
\end{itemize}
Indirect measurement methods of tool wear, such as force/torque, acceleration, temperature, motor currents, and acoustic sensors \cite{bleicherSensorActuatorIntegrated2023}, produce data that must be processed and analyzed using intelligent models to gain interpretable insights. In contrast, direct measurement techniques, including image recognition, optical measurements, and durability assessments, offer more straightforward data interpretation. The indirect method, or online TCM, requires no downtime and is therefore the preferred approach in many studies. Research in this area typically falls into two categories: physic-based models \cite{awasthiPhysicsbasedModelingInformationtheoretic2022} and data-driven models \cite{liIntelligentToolWear2022}. Physic-based models offer detailed theoretical explanations and mechanism analysis, such as monitoring force modeling coefficients during cutting to track tool wear. However, due to the complex non-linear nature of cutting processes, it is difficult to parameterize and some effects, such as cutting temperature or lubrication conditions, are ignored to simplify the model. This limits the accuracy of the model when there is no clear physical mechanism. As a subset of Industry 4.0, rapid advances in computing, digitization, and artificial intelligence (AI) have made it easier to implement data-driven models. These models use deep learning techniques, a subset of machine learning, to correlate sensor readings with tool wear, eliminating the need for a deep understanding of the physical processes. 
However, a fundamental understanding of sensor selection and input types remains necessary for an effective data-driven approach to tool wear prediction. Fig. \ref{fig:Sensors_and_wear} presents an overview of this essential knowledge and the sensors most commonly used for both direct and indirect measurement methods, with this study focusing on indirect measurements.
The Fig. \ref{fig:Sensors_and_wear} demonstrates how various types of wear, such as flank wear, lead to changes in the tool geometry (\autoref{fig:Sensors_and_wear}, (1)), which in turn affect the cutting forces. These forces can be categorized into static (mean) and dynamic components. As cutting forces directly interact with wear, they serve as primary indicators for TCM's. In milling, dynamic cutting forces play a particularly important role, as the intermittent engagement of cutting edges results in complex time-dependent interactions that can be effectively analyzed in the frequency domain. Direct measurement of cutting forces is possible through the use of a dynamometer, which is a widely used sensor in research. Nevertheless, the high cost and invasive installation of dynamometers limit their industrial application. A popular alternative are current sensors, which are often already integrated into machines. These sensors provide an indirect way of force measurement by correlating current consumption with the required cutting force. In a closed-loop system, dynamic forces are a contributing factor to vibrations, which are induced based on the process and machine dynamics (\autoref{fig:Sensors_and_wear}, (2)). These vibrations result in an uneven machined surface (\autoref{fig:Sensors_and_wear}, (3)), which subsequently influences the cutting thickness in the following tool engagement (\autoref{fig:Sensors_and_wear}, (4)). Consequently, the cutting thickness varies over time, further affecting the dynamic cutting forces. In this closed-loop system, the only indirect measurement method is vibrations, which can be measured with accelerometers. Accelerometers offer several advantages, including cost-effectiveness, minimal calibration requirements, and seamless integration into machines. The high sampling rates of accelerometers facilitate detailed time-domain analysis, capturing transient effects such as short peaks resulting from crater wear on the cutting edge. Moreover, accelerometers are particularly adept at frequency-domain analysis due to their capacity to cover a broad frequency range and their ability to be strategically positioned to minimize ambient noise. The utilization of advanced signal processing techniques, such as the Fast Fourier Transform (FFT), wavelet analysis, or the Welch method, ensures effective filtration of frequency background noise. In summary, accelerometers enable efficient and effective implementation of TCM's, which is why this study employs a data-driven approach using accelerometers.

\section{Related Work}

The paper employs a single acceleration signal to generate training data for data-driven models. This decision is motivated by the challenges associated with multi-sensor setups, such as the inability to integrate all sensors into every machine and the high costs or time requirements associated with sensor integration. These challenges hinder the generalization of results of multi-sensor setups across diverse applications.

To provide context for the single-sensor approach, we still introduce several multi-sensor systems (MSS) as a basis for comparison.  A multi-sensor system has the advantage of combining diverse sensor types and thus different signal inputs. In addition, the same sensor type can record data at different positions, typically near the tool, the workpiece, or the spindle to capture specific aspects of the milling process. One approach in multi-sensor systems is to focus on a single area, similar seen in simpler sensor setups.
In some studies, sensor placement focuses solely on the spindle, measuring spindle vibrations across multiple axes, spindle current, spindle force, and machine axis positions \cite{anDatadrivenModelMilling2020}. Other studies focus on the workpiece, using a stationary dynamometer under the workpiece (a "force plate")  with additional sensors for the monitoring of spindle current and vibration \cite{zhaoDeepLearningIts2019}.

However, most papers employ a multi-sensor approach by distributing sensors at various locations. These setups often use a force plate as the primary data source, supplemented by additional sensors for vibration and current measurements positioned elsewhere on the machine \cite{liNoninvasiveMillingForce2023}.
With multi-sensor setups, the data volume is inherently larger and can be further enhanced by adding a few handcrafted features for training data-driven models.  Various model architectures have been applied in MSS research, including random forest \cite{misalMillingToolWear2024}, support vector machines \cite{wangToolWearState2023}, simple neural networks \cite{bagriToolWearRemaining2021}, and convolutional neural networks (CNN) \cite{chengResearchMultisignalMilling2024,liNovelEnsembleDeep2022, martinez-arellanoToolWearClassification2019}. Among these, long short-term memory (LSTM) networks are frequently used due to their ability to handle time-series data \cite{pengIntelligentMonitoringMilling2024, wangMachineToolWear2024}. A common limitation across these MSS-based studies is the invasive machine modifications needed to install sensors, which often require specialized setups and time-consuming sensor installation and removal. Consequently, these studies are typically limited to testing on a single machine or using publicly available datasets \cite{3aa1-5e83-21, a.agoginoMillingDataset2007,chen2010PHMSociety2021} that were originally employed in a competitive context. Most public datasets still consist of data from only one machine.

In contrast to multi-sensor systems, the following studies employ minimalistic sensor setups, utilizing only one sensor or a few single-axis sensors positioned at the same location to simulate the functionality of a 2- or 3-component sensor. Sensors are strategically placed as close to the machining process as possible. Commonly, they are placed either near the workpiece or near the toolholder, though a placement near the spindle is occasionally used. When placed near the toolholder, the sensors are integrated directly into the toolholder or attached as adapters between the toolholder and the spindle. Typical sensor choices include 3-component dynamometers, such as the force plate \cite{kiousDetectionProcessApproach2010}, and accelerometers, as these offer valuable insight into the dynamics of processes and machines through frequency analysis.
Because simple sensor setups yield only one type of data, additional processing is often required to improve interpretability. This is achieved by supplementing the data with process parameters or generating handcrafted features. Common handcrafted features include statistical measures such as arithmetic mean, variance, skewness, and kurtosis \cite{liToolWearMonitoring2023}. Another popular approach involves data transformation techniques. For example, Zhang et al. apply wavelet packet decomposition to enrich data content, followed by an autoencoder for further processing \cite{zhangToolWearMonitoring2020}. Another study employs variable mode decomposition (VMD) to extract meaningful signal features \cite{yangNovelMultivariateCutting2022}.
Popular models for TCM with simple sensor setups include deep neural networks (DNN), convolutional neural networks (CNN) \cite{caoIntelligentMillingTool2019}, and long short-term memory (LSTM) networks \cite{wangMultidomainFeaturesFusion2023}. 
For 2D CNN's, data must be transformed into an image-like format, either by concatenating time-series features into a matrix or through frequency-time transformations such as the Continuous Wavelet Transform (CWT) or the Short Time Fourier Transform (STFT), which offer a time-frequency perspective on the data.

Overall, these studies demonstrate promising results within their specific use cases, showing that models can identify correlations between sensor measurements and tool wear. However, model validation has typically been limited to a single milling process, with only variations in the process parameters. As a result, while the models were proven capable of capturing wear-related patterns in this context, their ability to generalize or identify universal patterns in wear behavior, remains largely unexplored. Another limitation is that most studies conducted milling processes in a dry state to accelerate wear. In industrial applications, milling is often performed with a cooling lubricant, which affects tool wear differently. This discrepancy between experimental conditions and real-world practices makes it challenging to assess how well these research findings would translate to most industry applications. In this study, we will address the limitations observed in previous research. As a result, this study makes the following contributions:
\begin{itemize}
    \item Employment of a cost-effective data-driven approach for the rapid implementation of TCM's.
    \item Utilization of a single accelerometer sensor, which is straightforward to integrate into a range of machines and processes.
    \item Validating the model's generalization capabilities  and the necessity of retraining by performing wear prediction on two machines.
    \item Identification of the minimum amount of training data required, thereby reducing the burden of data collection.
    \item Performance analyisis of multiple model architectures, including a ConvNeXt model, LSTM, decision tree, and supper vector classifier (SVC).
\end{itemize}

\section{Architecturedesign for wear prediction}
\subsection{Problem description and Pipeline}
Data-driven models require large datasets to capture the non-linear relationships and random variations inherent in complex machining processes. In addition, these models are often tailored to the specific processes they are trained on. In real world applications, however, processes can frequently change, and for companies to justify the investment in wear detection systems, these models need to be adaptable to new or evolving processes. The complexity of tool wear makes it challenging to develop a generalizable data-driven model that can be applied across various processes, limiting the practical implementation of such research. 
To address this issue, this paper proposes a different approach by comparing various data-driven models and evaluating their performance based on the quantity of available training data. The training data is based on a single acceleration sensor, enabling fast and cost-effective model training for any process. It also makes it easier to adapt the model to new processes, making it more suitable for transfer learning.
The training datasets contain acceleration data of vibrations during the milling process. From these raw signal transformed entities, relevant features are extracted with the objective of predicting tool wear. The problem is framed as a binary classification task with the aim of determining whether wear with a negative impact is present.
This simplification aligns with industry needs, as companies primarily require clear indications of when wear starts to degrade machining quality, rather than a continuous regression-based wear estimation.  
\subsection{Data preprocessing}
\begin{figure*}[htbp]
    \centering
    \includegraphics[width = 1 \textwidth]{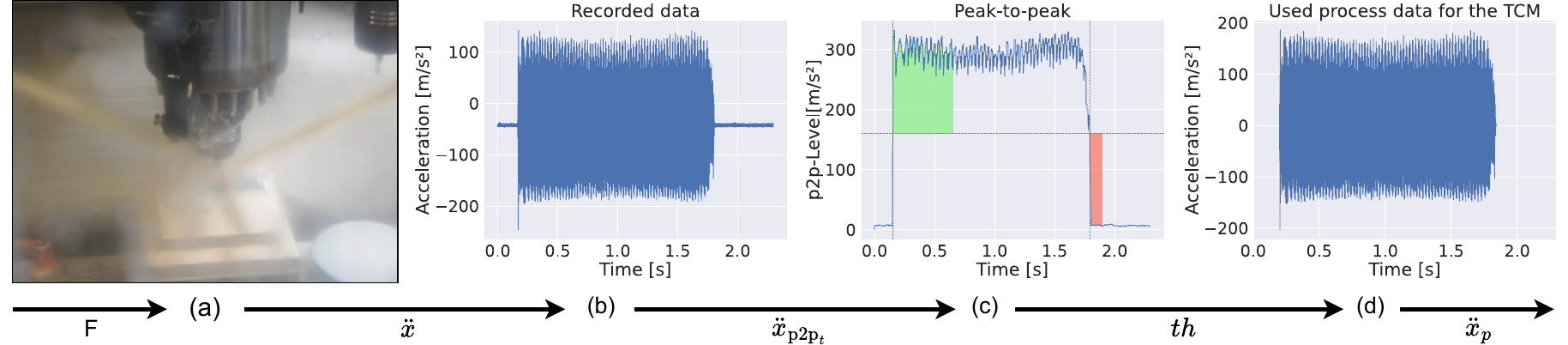}
    \caption{Pipeline for generating process data: (a) The milling force is converted into acceleration data \(\ddot{x}\). (b) A plot of the recorded acceleration data is shown. (c) The result of the p2p function \(\ddot{x}_{\text{p2p}_t}\) is displayed, along with the duration (green or red areas) where \(\ddot{x}\) exceeds or falls below the threshold  \(th\). Using this analysis, the acceleration data is segmented into process data (d). The resulting process data \(\ddot{x_p}\) is used for our classification pipeline.}
    \label{fig:Data_Input}
\end{figure*}

The accelerometer generates a continuous stream of acceleration data $ \ddot{x} $ that includes not only the critical process information related to tool wear, but also all the surrounding recordings, such as spindle run-up, tool change, and the transitional moments when the tool is starting or finishing its milling operation. To filter out these unwanted data, the signal is automatically processed by analyzing its peak-to-peak (\(\ddot{x}_{\text{p2p}_t}\)) over time steps \textit{t} with a moving window of length \textit{w}.
\begin{multline}
\ddot{x}_{\text{p2p}_t} = \max \left( \ddot{x}_{t - w + 1}, \ddot{x}_{t - w + 2}, \dots, \ddot{x}_t \right) \\
- \min \left( \ddot{x}_{t - w + 1}, \ddot{x}_{t - w + 2}, \dots, \ddot{x}_t \right)
\end{multline}
A threshold is calculated to isolate the segments of the signal that correspond to the actual milling process. For this, the \(\ddot{x}_{\text{p2p}_t}\) graph is sorted in ascending order ($\uparrow$) and then the average lowest value $m_{\text{p2p}}$ is determined as described below. Here \textit{N} is the length of the \(\ddot{x}_{\text{p2p}_t}\) curve.
\begin{align}
    m_{\text{p2p}} &= \frac{1}{ N} \sum_{i=\alpha_0\cdot N}^{\alpha_1\cdot N} \ddot{x}_{\uparrow\text{p2p}_i} , \\
    th &=  \alpha_2\cdot m_{\text{p2p}}
\end{align}
The $\alpha_{i}$ represent empirically determined values, with $\alpha_0 = 0.01$, $\alpha_1 = 0.03$, $\alpha_2 = 10$.
The signal is also trimmed according to the duration for which it is below or above the threshold $th$, thereby preventing the misclassification of brief anomalies with comparable vibration energy. The data preprocessing is explained in Fig. \ref{fig:Data_Input}.
\subsection{Feature engineering}
The acceleration data takes a measurement every 65 $\mu s $, giving a sampling rate of approximately 15,38 kHz. This leads to a Nyquist border with 7,89 kHz,
so the data is highly frequency sensitive and prone to short time-variant fluctuations or changes. But one minute of data already contains almost a million data points packed all in a one-dimensional acceleration signal. This complexity posing a challenge in identifying the complete range of characteristics and correlations related to wear. To reduce the complexity current approaches are data transformations and/or feature extraction. These approaches often reducing the amount of data while preserving the essential information.

Data or signal transformations are essential to the analysis of vibration signals, particularly in the frequency and time-frequency domains, as frequency characteristics provide crucial insight into the milling process. Traditional techniques such as the Fast Fourier Transform (FFT) are still widely used. But in milling, how frequencies change over time is also important, making time-frequency transformations more relevant. Common methods in this area include the Wavelet Transform, Hilbert Huang Transform, Empirical Mode Decomposition (EMD) and the Short-Time Fourier Transform (STFT). In this study, the STFT is used in combination with the Welch method to transform the data. The STFT provides time-resolved frequency analysis by applying the Fourier transform to short, overlapping segments of the signal. The Welch method further improves spectral estimation by averaging segments of the Fourier transform, improving frequency accuracy and reducing noise. The resulting transformation produces a two-dimensional, picture-like representation where the y-axis represents frequency \(f\) and the x-axis represents time \(t\). The formula for an STFT is as follows
\begin{align}
S[f, t] = \sum_{m=0}^{M-1} \ddot{x}_t[m] e^{-2j \pi f \frac{m}{M}}
\end{align}
In the STFT, \(\ddot{x}_t[m]\)  represents a windowed segment of the signal to which the Fourier transform is applied. A fundamental property of the STFT is its fixed frequency distribution for every application. This ensures that, irrespective of the signal's actual frequency content, the energy components are consistently mapped to the same frequency bins in the STFT representation. A further advantage of the STFT is that it does not increase the overall data volume when transforming a one-dimensional signal into a two-dimensional representation, which makes it computationally efficient, particularly when using convolution-based algorithms, which benefit from the structured nature of the transformed data. In contrast, other time-frequency transformations, such as the wavelet transform and empirical mode decomposition (EMD), introduce a significantly higher computational burden. The wavelet transform necessitates repeated convolutions with different basis functions, resulting in increased processing time, while EMD involves an iterative decomposition process that is inherently computationally expensive. These methods yield high-dimensional outputs that require substantial resources for further processing and possess high intrinsic computational costs, rendering them challenging to implement in real-time industrial applications. To further reduce noise and data complexity, this study employs the Welch method in combination with the STFT. Fig. \ref{fig:Dif_models}a illustrates the different data transformation approaches, with the corresponding transformations shown on the left. The transformed data can now be used as input for a data-driven model, allowing it to perform feature extraction automatically. Alternatively, manually engineered features can be generated using custom algorithms to further reduce the input dimensionality before feeding the data into a classification model. In this study, both approaches are explored.

For the purpose of wear detection in milling and vibration signal analysis, well-established features have been selected. Statistical features remain fundamental and include metrics such as mean, peak-to-peak amplitude, kurtosis, and skewness. In addition to these, more specialized features are considered to capture changes in signal complexity, which often indicate wear. In the frequency domain, an increase in signal complexity is characterized by the presence of multiple frequency components, which is quantified using the Spectral Entropy feature, which measures how evenly energy is distributed across different frequencies. In the time domain, disorder can be assessed using the Fractal Dimension, which describes the irregularity of vibration patterns. Another important feature extraction method involves separating the signal's energy in the frequency domain into periodic and aperiodic components. An increase in aperiodic energy, i.e. energy from frequencies not associated with fundamental process frequencies, such as the tooth entry frequency, can be an indicator of tool wear. The features employed in this study, along with their respective descriptions, are outlined in Tab. \ref{table:features_1}, \ref{table:features_2}.

\subsection{Different models}

\begin{figure*}[htbp]
    \centering
    \includegraphics[width = 0.9 \textwidth]{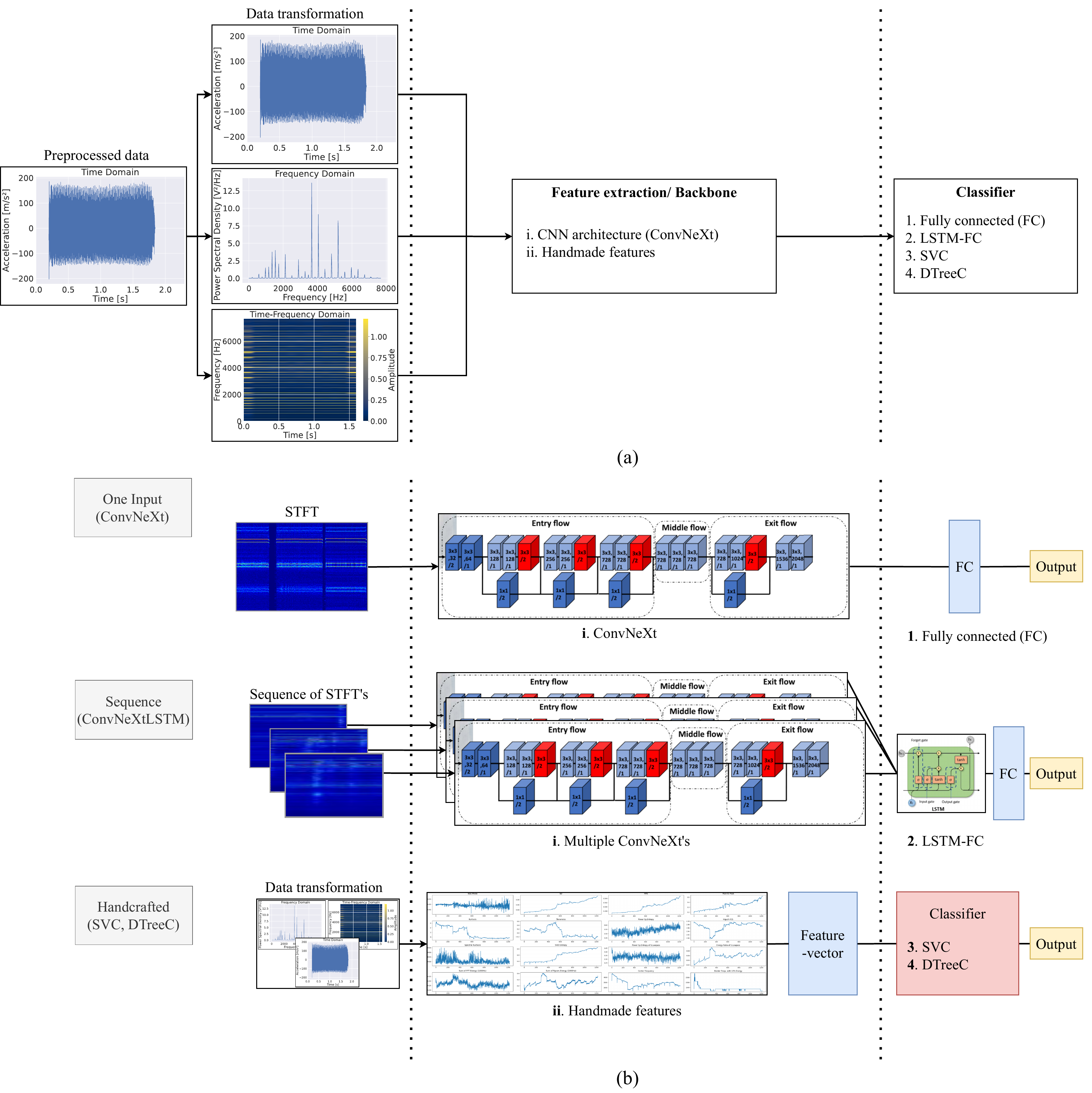}
    \caption{(a) The diagram illustrates the transformation of acceleration data into different domains and the subsequently extraction of features using either a CNN architecture (i) or proprietary algorithms (ii). These features are then utilized by various classifiers (1, 2, 3, 4) to predict wear. (b) The architectures of the four classifiers used in this paper, categorised by their input type, are shown. Images from the following sources were used to create this diagram: \cite{westphal_machine_2021,rungeReviewDeepLearning2021}}
    \label{fig:Dif_models}
\end{figure*}

One objective of this research is to identify the most suitable model for an industry-ready single-sensor installation for tool wear detection. To this end, a range of state-of-the-art architectures are evaluated, differing in complexity and computational cost. The aim is to determine the level of model intelligence required for effective wear prediction. Fig. \ref{fig:Dif_models}a illustrates the structures underlying the different models. The modeling process consists of two key components: feature extraction from the signal (\autoref{fig:Dif_models}, (i, ii)) and classification (\autoref{fig:Dif_models}, (1,2,3,4)). The first feature extraction approach (\autoref{fig:Dif_models}, (i)) relies on a model-based backbone, which is computationally more intensive but allows the model to learn relevant features autonomously. As discussed in the previous chapter, the STFT is used as the input representation. The task is then framed as an image classification problem, where the model must recognize wear conditions from the transformed data. At present, two leading architectures dominate image classification: Convolutional Neural Networks (CNN) \cite{ krizhevskyImageNetClassificationDeep2017, simonyanVeryDeepConvolutional2015a} and Vision Transformers (ViTs)\cite{dosovitskiy_image_2021}. While ViTs have demonstrated strong performance in the field of computer vision, CNNs remain superior in many domains, particularly when training data is limited and computational efficiency is crucial \cite{zhuUnderstandingWhyViT2023}. Recent advances in deep learning have led to hybrid CNN-ViT architectures \cite{khanSurveyVisionTransformers2023}, combining the strengths of both approaches \cite{yunusaExploringSynergiesHybrid2024}. Among these, CNN-based models continue to be the dominant component. In this study, the CNN-ViT hybrid model, ConvNeXt, is selected as the backbone. ConvNeXt builds on the widely used ResNet framework \cite{he_deep_2015}, incorporating architectural enhancements inspired by the Swin Transformer \cite{liu_swin_2021}. Benchmark studies show that ConvNeXt performs on par with or better than Vision Transformer in many cases \cite{liu_convnet_2022}.

As demonstrated in Fig. \ref{fig:Dif_models}b, the first two models utilize ConvNeXt as the backbone.
\begin{enumerate}
    \item Model "ConvNeXt": Employs a simple fully connected layer as the classifier. While this model is computationally efficient, it processes STFTs independently and is unable to account for temporal dependencies beyond the information embedded in each individual STFT.
    \item Model "ConvNeXtLSTM": Designed to handle sequential wear data, and processes a sequence of STFTs. Each STFT is first passed through the ConvNeXt backbone, generating a sequence of features. These features are then processed by an LSTM-based classifier with three layers (64, 32, and 32 LSTM units, respectively), followed by a fully connected layer. To mitigate overfitting, dropout (10\%) is applied to the first two LSTM layers.
\end{enumerate}
The third and fourth models rely on manually engineered features (Fig. \ref{fig:Dif_models}b, (ii)). In this approach, a moving window is employed to extract 20 handcrafted features at each step. These extracted features are subsequently fed into classical machine learning classifiers:
\begin{enumerate}
    \setcounter{enumi}{2}
    \item Model "SVC": A Support Vector Classifier with a radial basis function (RBF) kernel.
    \item Model "DTreeC": A Decision Tree Classifier using the Gini impurity criterion, without depth limitation.
\end{enumerate}

\subsection{Training and Evaluation process}
\begin{figure*}[htbp]
    \centering
    \includegraphics[width = 0.9 \textwidth]{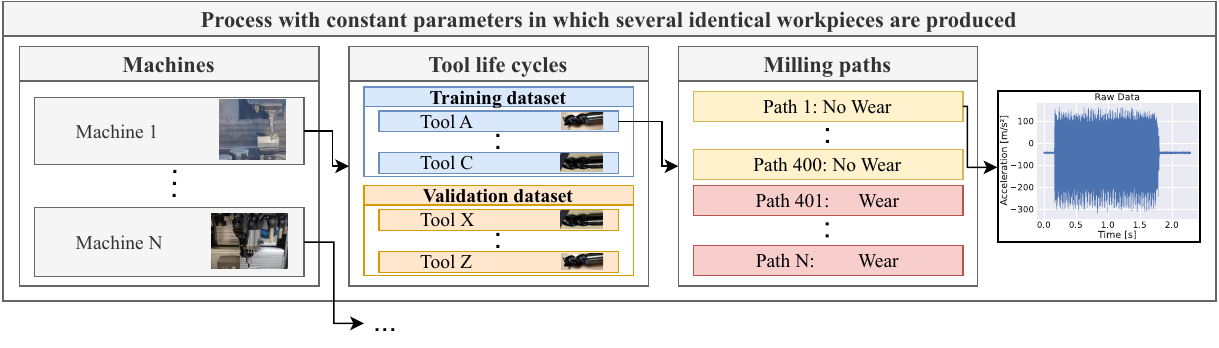}
    \caption{Structure of the datasets used for the models: Each process dataset includes multiple machines, and each machine dataset contains several tool life cycles. A single tool life cycle represents the use of one tool until it becomes worn. Within each dataset, multiple workpieces are produced.}
    \label{fig:Dataident}
\end{figure*}

The training data is organized in such a way that multiple process runs are performed for each milling operation. Each run tracking the entire tool life cycle of a tool: from completely new to completely worn out. During model training, it is critical to avoid random splitting of the data between training and validation sets, as this can lead to mixing of process runs. This is because each run contains a new tool with different wear behavior, making random splitting inappropriate.
To ensure a valid model evaluation, no part of a specific process run should be used in both training and validation. If the model is exposed to segments of the same run in both phases, it may overfit to the specific wear patterns of that particular run rather than learning to generalize across different tools and wear conditions. Fig. \ref{fig:Dataident} shows how the data is sorted in training data and validation data.

Two milling machines are utilized for the training and evaluation process. The first machine generates the training and validation data for the various models. It has a smaller working area and is therefore more resistant.  To assess model performance with minimal training data, we iteratively train each model with a decreasing number of process runs. This method allows us to evaluate how the model behaves with both rich and limited data, revealing which models are more robust with larger datasets and which perform better under data-constrained conditions. We also determine the minimum amount of training data required to achieve acceptable performance and the amount necessary for optimal results.
The next step is to assess how effectively the models trained on one process generalize to a different, yet similar, process. For this purpose, the milling machine is changed. This second machine is optimized for multifunctional machining but exhibits lower stability.  It is important to acknowledge that the two machines used in this study exhibit fundamentally distinct dynamic behaviors. Throughout the evaluation, the models are compared to determine their effectiveness in handling new unseen processes and to detect their relative strengths and performance differences.

\section{Experimental results}
\subsection{Setup}
Typical components used in a standard milling process were selected. For this experiment, the tool life cycle of a solid carbide square end mill with four flutes and a diameter of 12 mm is analyzed on two different machines: the first machine is the \textit{CHIRON FZ 15 S} and the second machine is the \textit{DMU 60 FD duoBLOCK}. The workpiece material is hardened steel (42CrMo4, +QT), chosen for its demanding properties.

\begin{figure}[htbp]
    \centering
    \includegraphics[width = 1 \columnwidth]{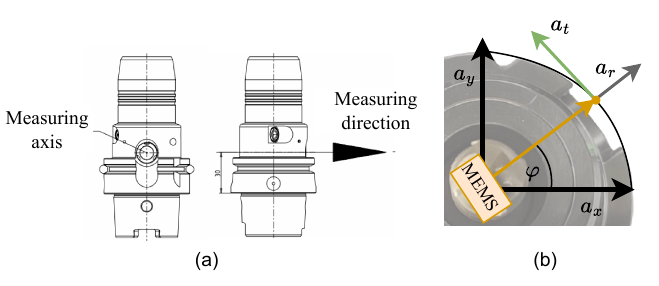}
    \caption{(a) Side view of the toolholder with the integrated 1-axis MEMS sensor. (b) Top-down view of the toolholder, illustrating how the sensor measures acceleration in the \(xy\)-plane, depending on the rotation angle \(\varphi\).}
    \label{fig:kinModel}
\end{figure}

As discussed previously, an accelerometer was selected due to its excellent balance of cost efficiency, ease of integration, and high-quality information for tool wear monitoring. To further enhance industrial application, this study employs a single-sensor configuration to minimize costs while maximizing integrability. However, using only one sensor introduces challenges that must be addressed through optimal sensor selection. Since accelerometers excel in capturing frequency-domain information, several key factors must be considered to ensure reliable wear detection:
\begin{itemize}
    \item High sampling rate to accurately capture rapid signal variations.
    \item Broad frequency range and high frequency resolution to detect relevant vibration signatures.
    \item High resonance frequency to avoid distortions and ensure accurate measurements.
    \item Extremely low noise density to maintain signal clarity.
    \item Positioned close to the cutting process to maximize sensitivity to process-related vibrations while minimizing external noise interference.
\end{itemize}
To meet these requirements, a tool holder-integrated sensor was selected, allowing direct measurement as close to the cutting process as possible while simplifying installation.
Following a thorough evaluation, the iTENDO² from Schunk  was identified as the optimal solution (Fig. \ref{fig:kinModel}a). The sensor used is a single-axis, low noise, high frequency MEMS acceleration sensor that measures radial 
acceleration ($a_r$). As the sensor is placed on the rotation axis of the toolholder, it is able to effectively measure the acceleration on the x-axis ($a_x$) and y-axis ($a_y$) over time, as it rotates with the toolholder and spindle speed.  Fig. \ref{fig:kinModel}b illustrates the measurement behavior, which presents a top view of the iTENDO² in the xy plane with the axis of rotation being the z-axis.
A rotation matrix is applied to calculate $a_r$, accounting for the changing contributions of $a_x$  and $a_y$ based on the rotation angle \(\varphi\).:
 \begin{align}
    \left[\begin{array}{l}
        a_{r} \\
        a_{t}
        \end{array}\right]=\left[\begin{array}{c}
        \cos (\varphi) a_x+\sin (\varphi) a_y \\
        \cos (\varphi) a_y-\sin (\varphi) a_x \label{eq:drehM}
    \end{array}\right]
\end{align}
If the sensor is installed optimally, it does not measure tangential acceleration \(a_t\). As a result, the measured accelerations are dynamically mixed depending on the angle of rotation, enriching the signal with more detailed acceleration information. However, this also makes the signal more challenging to interpret. The reason for this is that the rotation matrix multiplies frequencies in the time domain, as both $a_x$ and $a_y$ also consist of frequency components. In accordance with the convolution theorem, the frequencies in the frequency domain are also convolved.
This particular sensor/toolholder combination is still chosen because the sensor's position allows it to capture undamped process vibrations more effectively. In addition, the setup can be easily transferred to other machines and processes, requiring only replacement of the toolholder to record training data under similar conditions. This modularity aligns seamlessly with the key contribution of this research: the development of a novel TCM pipeline that enables models to be swiftly adapted to a range of industrial milling processes for wear classification.
\subsection{Design of Experiment}
\begin{figure}[htbp]
    \centering
    \includegraphics[width = 1 \columnwidth]{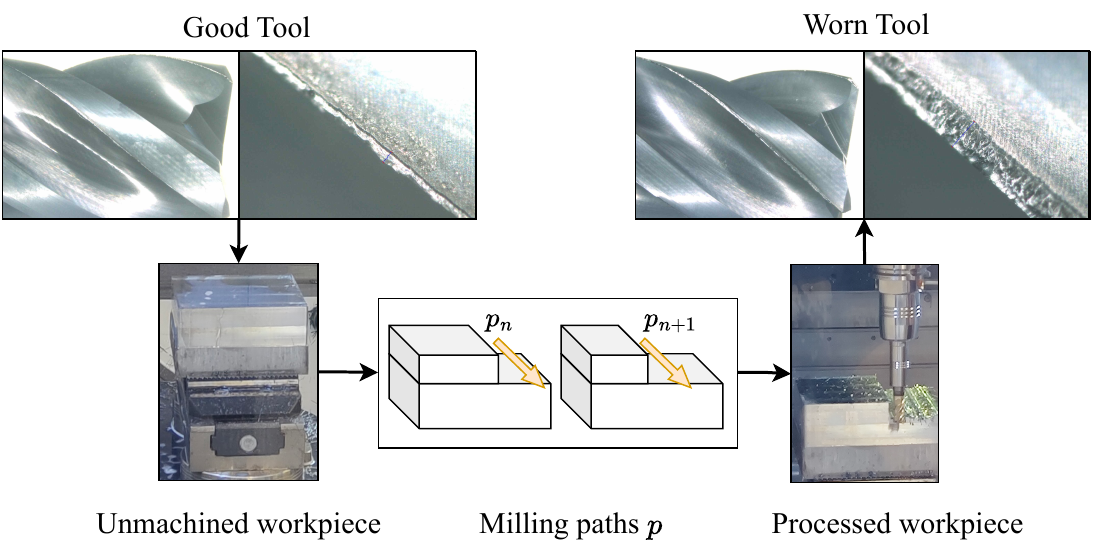}
    \caption{Procedure for one tool life cycle, involving machining N milling paths until the tool becomes excessively worn.}
    \label{fig:Experiment}
\end{figure}

\begin{figure}[htbp]
    \centering
    \includegraphics[width = 1 \columnwidth]{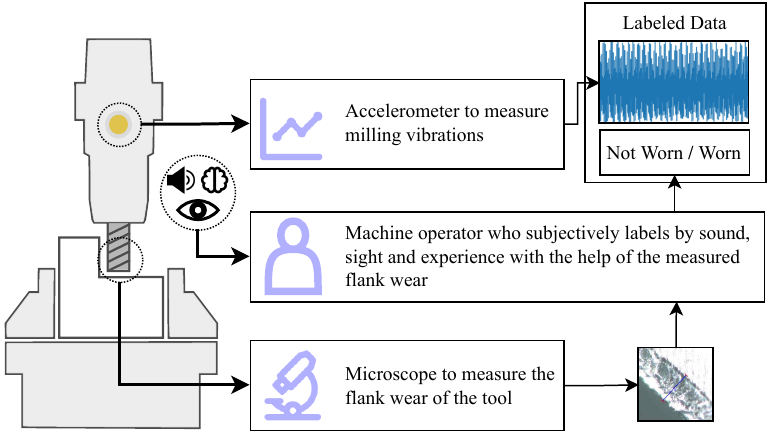}
    \caption{Overview of the tool state labeling process. Here the ToolState label is "Not worn" or "Worn"}
  
    \label{fig:how_label_data}
\end{figure}

In this experiment, a flat surface is created on a metal block using side milling. This milling process is repeated continuously until the cutting tool shows significant wear. The milling operation is a down-milling approach, with coolant lubricant applied throughout the process. Milling is performed exclusively along the xy plane. Detailed process parameters are given in Tab. \ref{table:process_param} and the visualization of the experiment is shown in Fig. \ref{fig:Experiment}.
In order to accurately characterize the acceleration data, the flank wear of the end mill is measured at regular intervals. When the flank wear reached 0.2 mm, the tool is considered to be highly worn, indicating the completion of one tool life cycle and the process run restarts with a new tool.
Flank wear is a reliable indicator of tool condition, directly influencing cutting performance. However, it is important to note that flank wear is just one of several wear mechanisms and its reliability as a universal wear metric is dependent on the specific milling conditions. This variability in performance makes it impossible to define a strict threshold for tool replacement based solely on flank wear measurements. Instead, an approximation is used to determine wear severity. Since this study aims to develop an industry-oriented solution, the model is not designed to predict flank wear itself but to classify the tool's condition as either usable or worn. To enhance the labeling process, an experienced machine operator manually evaluates the tool's condition and assigns a ToolState label, as illustrated in Fig. \ref{fig:how_label_data}.
The evaluation is based on multiple wear indicators, including workpiece surface quality, crater formation on the tool, chip shape and color, and audible changes in cutting noise. However, this labeling process is inherently subjective, as tool wear progression is complex and difficult to pinpoint with absolute temporal precision.  Consequently, it is not possible to objectively determine the exact moment when wear begins to significantly degrade the milling process. This means that no direct accuracy or quality metric can be established to evaluate the data sets.  Overall, the ToolState label is the most suitable option for wear prediction in this research.
\section{Evaluation}
\begin{figure}[htbp]
    \centering
    \includegraphics[width = 1 \columnwidth]{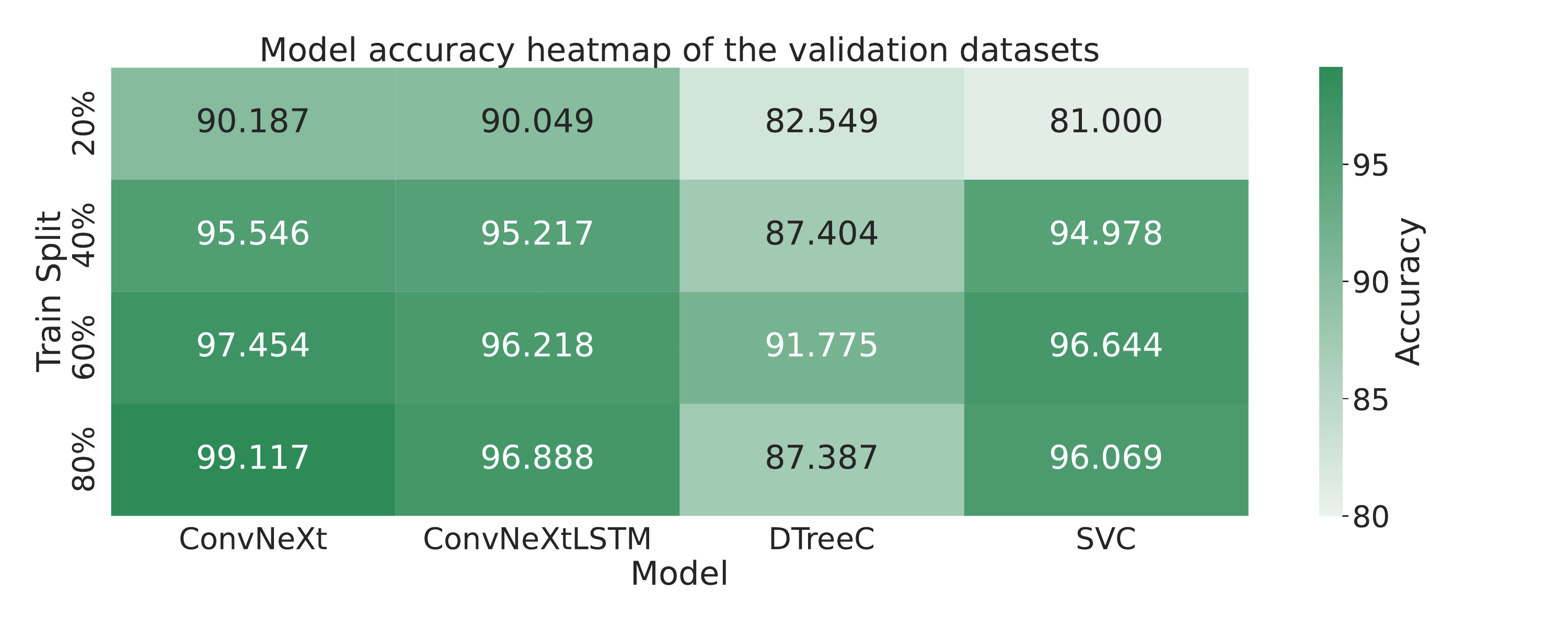}
    \caption{A heatmap showing the accuracy of various trained models on the validation datasets. }
    \label{fig:heatmap_val1}
\end{figure}

Each model is trained and validated using the Chiron machine dataset. The ratio of training data to validation data is divided into four steps. It starts with 20\% training to 80\% validation and changes the ratio in steps of 20\% up to the ratio of 80\% training to 20\% validation. A total of 30 models per architecture are trained across all four splits. Each model is trained over a maximum of 16 epochs, with often only a slight improvement in model performance observed from epoch 12 onward. The model weights that performed best on the validation dataset are used. Fig. \ref{fig:heatmap_val1} shows the accuracy of each model architecture as a heatmap over the amount of training data.
It can be seen that there are only small differences between the training data of 80\% and 60\%. This indicates that a good prediction of machine wear can be achieved by training on 60\% training data, which corresponds to three tool life cycles.

The analysis of model accuracy reveals that ConvNeXt-based models outperform classifiers that utilize manually designed features. It seems that the ConvNeXt backbone is more effective in extracting features that capture the wear behavior of the milling process. The "DTreeC" and the "SVC"-Model, may have limited representation capabilities, which restricts their ability to model complex wear patterns. Another key aspect influencing model performance is signal noise, particularly due to the sensor's rotation. The CNN-ViT hybrid architecture handles this challenge more effectively by focusing more on spatial patterns rather than relying on exact values. Despite this, feature-based classifiers have significantly fewer parameters, making them computationally efficient while still delivering solid performance. Interestingly, the ConvNeXt model with a simple dense layer as a classifier outperforms its LSTM-enhanced counterpart. This suggests that the STFTs already contain sufficient temporal information, making additional sequence modeling unnecessary. Furthermore, short-term variations, such as sudden shocks from a cutting edge, appear to carry more predictive value than long-term trends, given that wear generally increases gradually. Another possible explanation is that LSTMs tend to overfit, especially when trained on a relatively small dataset, highlighting the risk of excessive model complexity without a proportional gain in performance.

\begin{figure*}[htbp]
    \centering
    \includegraphics[width = 1 \textwidth]{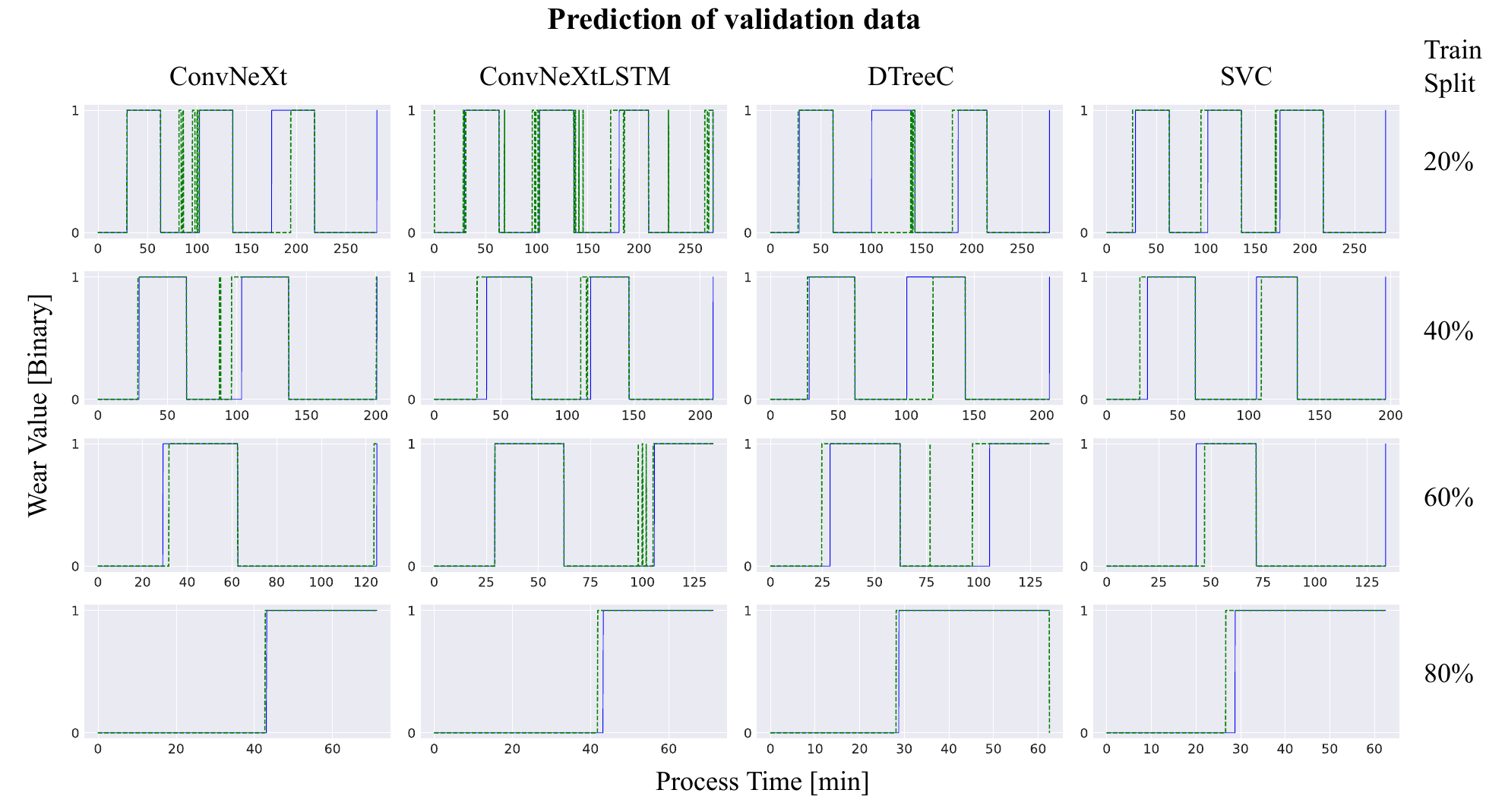}
    \caption{The time-based wear predictions of the various models, which have been trained on different amounts of data. The dashed green line represents the model's predictions, while the blue line indicates the true labels.}
    \label{fig:pred_plots1}
\end{figure*}

Although the prediction accuracy or binary accuracy reflects the overall quality of the model, in reality it is more important how the model predicts wear over time, in particular when the model changes its prediction from not worn to worn. Fig. \ref{fig:pred_plots1} shows the prediction of wear over time. The y-axis illustrates the categories of the binary classification, which are defined as follows:
\begin{equation}
   C: \{0, 1\} \to \{\text{Not worn}, \text{Worn}\} 
\end{equation}
The wear subjectively labeled by the operator (true label) is shown in blue and the model prediction is shown in green. Depending on the plot, several tool life cycles are shown concatenated in one plot. When the label changes from Worn to Not worn, a new tool life cycle begins. For post-processing, a simple filter is applied to the classification results to address the logical inconsistency that a tool cannot briefly appear worn and then revert to an unworn state. This filter eliminates a small proportion of such outliers from the plots, thereby enhancing both the accuracy over time and the readability of the visual representations. However, prominent outliers remain visible in the plots.  
Figure \ref{fig:pred_plots1} illustrates that time accuracy improves as the size of the training set increases. It is noteworthy that certain models, such as "ConvNeXt", demonstrate relatively good time accuracy even with only 40\% of the training data. The "SVC", despite its simplicity, achieves only slightly lower time accuracy compared to more advanced models.
The results together demonstrate that a minimum of two, preferably three, tool life cycles is required to train a model with the capacity to make reliable predictions. This assumption is based on the premise that the transition from no wear to worn occurs with minimal time deviations. In practice, recognizing wear too late is undesirable. To mitigate this, the transition from "not worn" to "worn" could be adjusted slightly earlier in the labeling process, thereby introducing a calculated bias for the early prediction of wear. Alternatively, if avoiding such a bias is preferred and high time accuracy is still desired, training on data from four or more tool life cycles becomes necessary.

\begin{figure}[htbp]
    \centering
    \includegraphics[width = 1 \columnwidth]{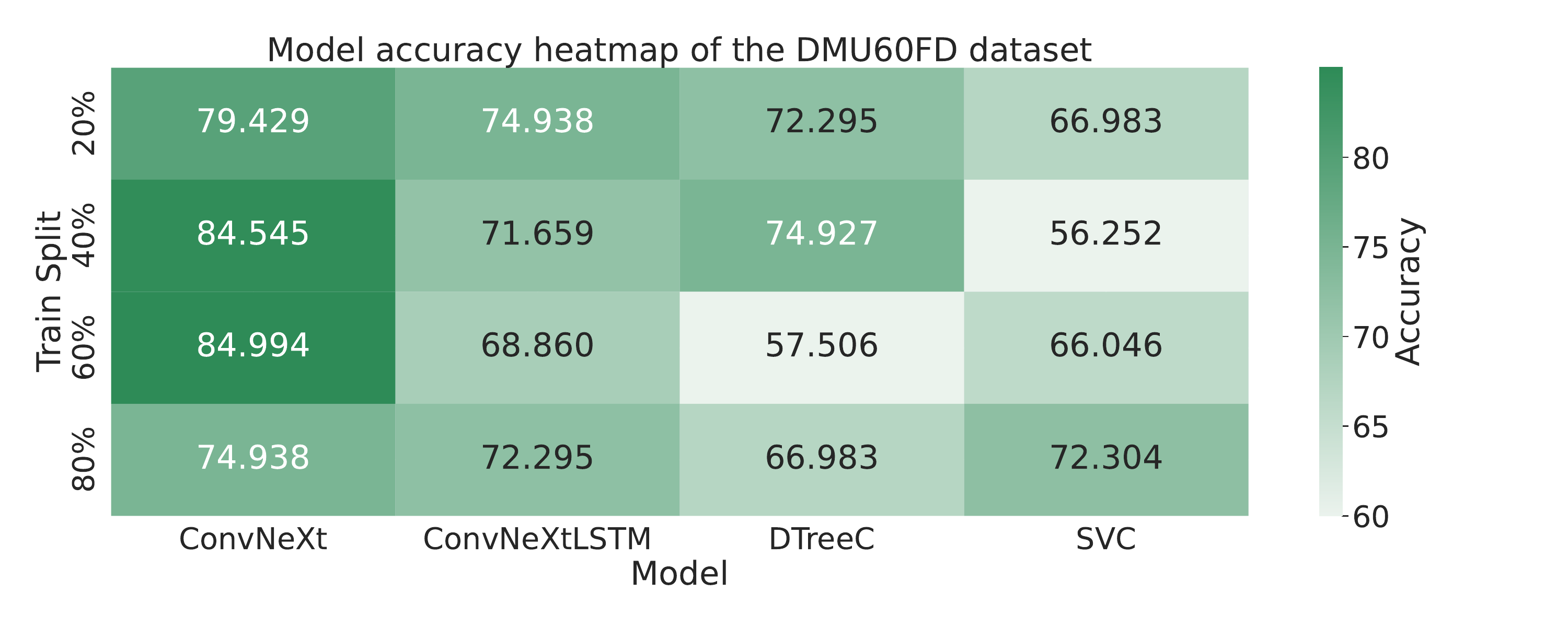}
    \caption{A heatmap showing the accuracy of the different trained models on the DMU60FD dataset.}
    \label{fig:heatmap_val_dmu}
\end{figure}

\begin{figure*}[htbp]
    \centering
    \includegraphics[width = 1 \textwidth]{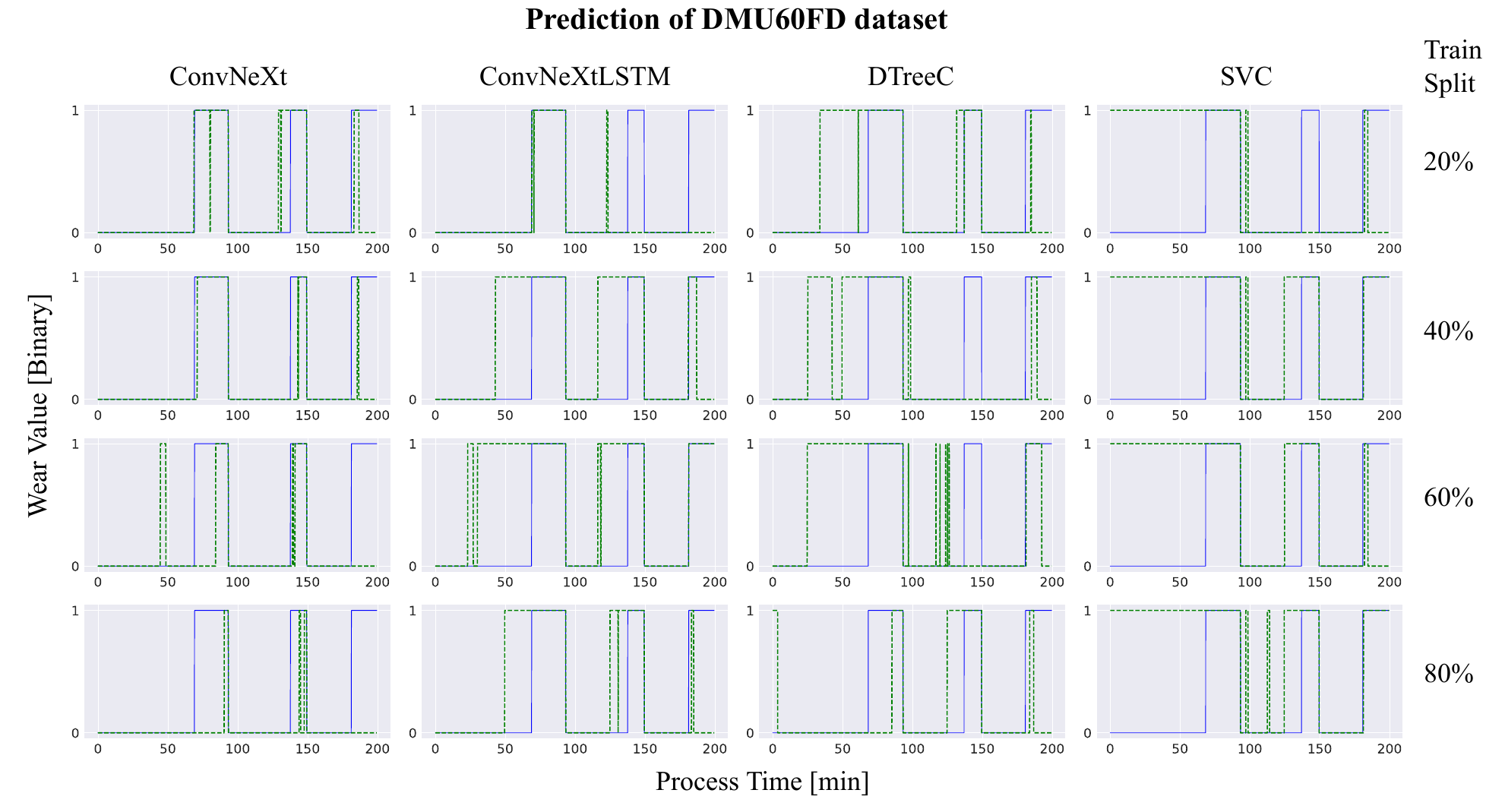}
    \caption{The time-based wear predictions of the DMU60FD dataset. The dashed green line represents the model's predictions, while the blue line indicates the true labels.}
    \label{fig:TimePlot_dmu}
\end{figure*}

\begin{figure}[htbp]
    \centering
    \includegraphics[width = 1 \columnwidth]{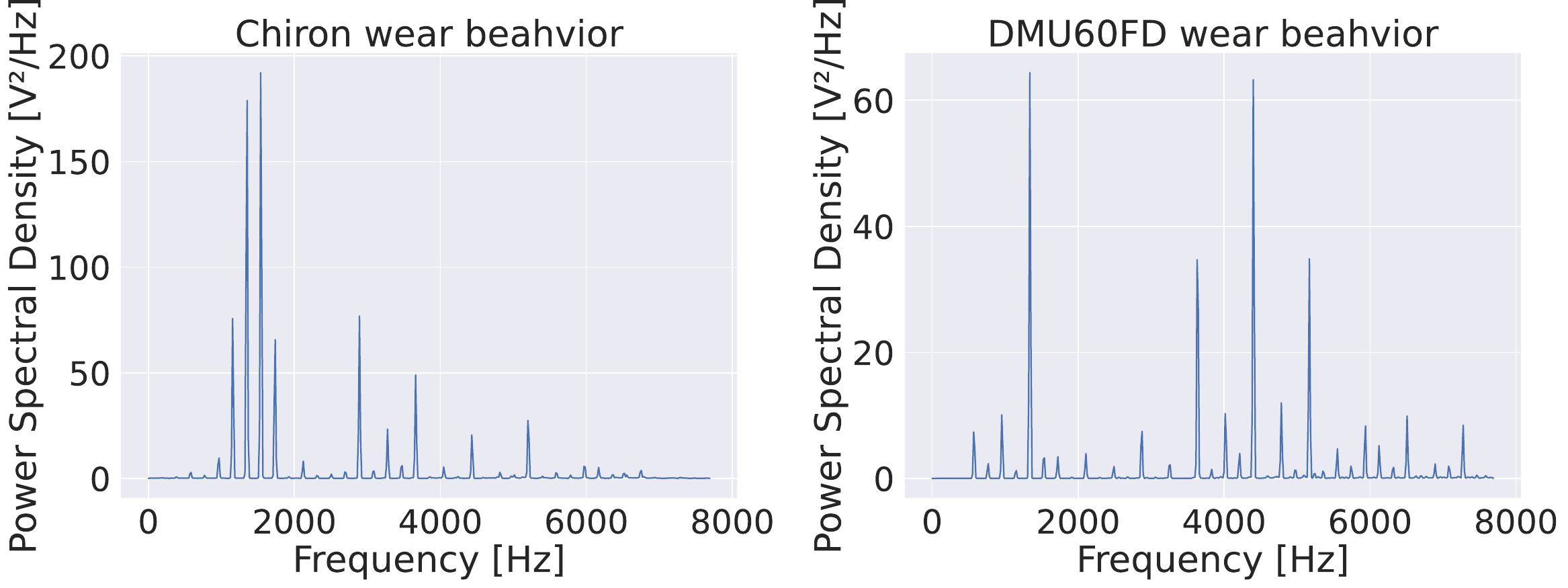}
    \caption{Display of two FFTs with a worn tool, each from the Chiron and DMU60FD}
    \label{fig:FFT_vgl}
\end{figure}

These observations hold true when the models are applied to an identical process with the same machine. The next step is to examine the behavior of the trained models in relation to data from a different machine. For this, Fig. \ref{fig:heatmap_val_dmu} illustrates the accuracy of the various model architectures, using a dataset from the DMU60FD machine.
It should be noted that the training data of the different splits are identical to those used in the heatmap in Fig. \ref{fig:heatmap_val1}.  It is evident that the models on the unseen machine exhibit reduced performance, yet the ConvNeXt model continues to demonstrate the most optimal outcomes.  Interestingly, models trained with the largest dataset (80\% training data) perform worse than those trained with smaller datasets. This suggests that the change of a machine has fundamentally altered the wear behavior, causing models trained on extensive data to overfit to the specific characteristics of the original machine. Models trained with 40\% of the training data achieve the highest accuracy. An examination of the accuracy over time (Fig. \ref{fig:TimePlot_dmu}) reveals that the models are capable of detecting changes in the DMU60FD dataset and successfully predict wear in some tool life cycles. 
Nevertheless, when considering all models collectively, while each tool life cycle is successfully recognized by at least one model, no single model achieves consistent accuracy across all cycles. This suggests that the findings derived from the training dataset can be applied to the DMU60FD dataset.  However, no overarching patterns are identified that apply to all tool life cycles. 
This discrepancy may be attributed to the differing wear behavior in the frequency domain between the two machines, as illustrated in \autoref{fig:FFT_vgl}. For the Chiron, significant changes occur primarily below 4000 Hz, whereas in the DMU60FD, wear-related variations are also observed above 4000 Hz. It is possible that the models focus predominantly on the sub-4000 Hz range, where the DMU60FD exhibits only partial changes. Consequently, the models may encounter difficulties in fully capturing wear patterns on the DMU60FD, as the most relevant wear-related information for this machine may lie above 4000 Hz.

In summary, the fundamental issue associated with TCM's is also clear in this study, namely that the complex characteristics of each machine make it difficult to implement one prediction model across other unseen machines.  
To improve predictions, the models require additional training data from the specific machine.

\section{Conclusion} %
This paper addresses the question of how to develop a TCM system that can be transferred to industrial applications. For this, a modular approach and a non-invasive setup is employed so that a TCM can be installed for different milling processes with minimal effort. For a non-invasive setup, a 1D acceleration sensor is used, and for modality the sensor is integrated into a toolholder. In this research, the acceleration signal is processed in both the time domain and the frequency domain, to generate valuable data input for wear prediction or TCM. The best performing TCM approach uses short-time Fourier transform as input and automatic feature extraction through a transformer-inspired CNN architecture to predict wear.
 Given that the development of a TCM is more cost-effective when fewer training data are required, the study also investigates the amount of training data needed from a milling process to get a sufficient accuracy: To achieve the highest degree of accuracy in wear prediction, at least four tool life cycles should be included in the training data, two or three cycles for an acceptable accuracy. Among the models tested, the ConvNeXt model delivered the best results with 99\% accuracy.
Additionally, the models capacity to predict wear for an untrained milling machine is evaluated. Although models demonstrate some ability to transfer information and predict wear for certain tools, they struggle to fully capture the inherent complexity and variability of new milling machines, resulting in inconsistent wear predictions. This underscores the necessity for transfer learning to adapt the model to different machining conditions.
Concurrent with the publication of this article, ongoing research is investigating the amount of data required for effective transfer learning, determining which models adapt best, and refining training weight adjustments. Additionally, efforts are underway to explore efficient methods for industrial data collection and labeling to facilitate real-world deployment.

\section*{Data availability} 
The authors do not have permission to share data. 
\section*{Acknowledgements}
The authors would like to thank SCHUNK SE \& Co. KG for their valuable support.
\newpage
\onecolumn
\appendices
\clearpage
\section{Tables of this research}
\label{app1}

\begin{table*}[htbp]
\centering
\renewcommand{\arraystretch}{1.3}
\resizebox{\textwidth}{!}{%
\begin{tabular}{@{}p{0.35\textwidth}p{0.7\textwidth}@{}}
\toprule
\textbf{Feature Name}                     & \textbf{Formula / Description}                                                                                    \\ \midrule
Mean                                      & $\frac{1}{N}\sum_{i=1}^N x_i$, remains stable in an undisturbed process but shifts with wear \cite{liIntelligentToolWear2022}.  \\                    
RMS (Root Mean Square)                    & $\sqrt{\frac{1}{N} \sum_{i=1}^N x_i^2}$, measures signal energy, higher values indicate increased wear \cite{chenFeatureExtractionUsing2019}.  \\                
Standard Deviation (SD)                   & $\sqrt{\frac{1}{N-1} \sum_{i=1}^N (x_i - \bar{x})^2}$, quantifies signal variation and fluctuation intensity \cite{zhuToolWearCondition2021}.  \\       
Crest Factor                              & $\frac{\max |x|}{\text{RMS}}$, detects impact forces and transient events in the signal \cite{qinToolWearMonitoring2025}.  \\              
Kurtosis                                  & $\frac{1}{N} \sum_{i=1}^N \left( \frac{x_i - \bar{x}}{\text{SD}} \right)^4$, measures impulsiveness and sharp peaks \cite{wangMachiningQualityPrediction2024}.  \\          
Skewness                                  & $\frac{1}{N} \sum_{i=1}^N \left( \frac{x_i - \bar{x}}{\text{SD}} \right)^3$, indicates asymmetry in the signal distribution \cite{duboustMachiningCarbonFibre2016}.  \\         
Statistical Mode                          & $\text{Mode}(X) = \arg\max_{x \in X} f_{cnt}(x)$, identifies the most frequent value, robust to outliers.  \\         
SD of Statistical Mode                    & Measures the variability of the mode, indicating process stability.  \\                         
Peak-to-Peak (P2P)                        & $\max(x) - \min(x)$, tracks extreme signal values, sensitive to transient events \cite{chengDatadrivenOnlineDetection2021}.  \\   
Center Frequency                          & $\frac{\sum f_i*A_i}{\sum A_i}$, average of the spectral content, shows dynamic changes \cite{yanToolWearMonitoring2021} \\  
Dominant Frequency                        & $\arg\max_{f}(PSD(f))$, the frequency with the highest energy, can shift due to wear \cite{kurekToolWearClassification2024} \\  
\bottomrule
\end{tabular}%
}
\caption{Statistical Features and corresponding formulas. The sources refer to research papers that have used a comparable feature.}
\label{table:features_1}
\end{table*}

\begin{table*}[htbp]
\centering
\renewcommand{\arraystretch}{1.3}
\resizebox{\textwidth}{!}{%
\begin{tabular}{@{}p{0.35\textwidth}p{0.7\textwidth}@{}}
\toprule
\textbf{Feature Name}                     & \textbf{Formula / Description}                                                                                    \\ \midrule
Energy of Power Spectral Density (PSD)    & $\frac{\sum{|X(f)|^2}}{N}$, where $X(f)$ is the PSD of $x(t)$, reflects vibration energy and wear progression \cite{wangDigitalTwinsHighperformance2024}.  \\        
Power Spectral Entropy                    & $- \sum_{n=1}^{N} P(n) \log_2 P(n)$, where $P(n)$ is the probability distribution, indicating frequency spread due to wear \cite{jiangIdentificationMethodUnsteady2023}.  \\  
Periodic Energy                           & $\sum_{f \in F_{\text{process}}} E_f$, energy in harmonic components (e.g., spindle, tooth-pass frequencies)  \cite{liToolWearClassification2023}.  \\     
Aperiodic Energy                          & $E_{\text{total}} - \sum_{f \in F_{\text{process}}} E_f$, quantifies energy from non-harmonic, "chaotic" components (e.g., friction).  \\      
Relative Aperiodic Energy                 & $\frac{E_{\text{ap}}}{E_{\text{total}}}$, ratio of aperiodic energy to total energy, indicating noise and disturbances.  \\      
Auto-Correlation                          & $\frac{\sum_{i=1}^{N-\tau} (x_i - \bar{x})(x_{i+\tau} - \bar{x})}{\sum_{i=1}^N (x_i - \bar{x})^2}$, signal randomness and periodicity \cite{ostaseviciusMachineLearningApproach2021}.  \\      
Higuchi Fractal Dimension                 & Estimated via linear regression on $\{(\log(1/k), \log(L(k)))\}$, where $k$ is the scale parameter and $L(k)$ the curve length. Detects time series complexity \cite{liuIntelligentRobustMilling2017}.  \\      
\bottomrule
\end{tabular}%
}
\caption{Description of advanced features, mostly for the frequency domain. The sources refer to research papers that have used a comparable feature.}
\label{table:features_2}
\end{table*}

\begin{table*}[htbp]
\centering
\renewcommand{\arraystretch}{1.3}
\resizebox{\textwidth}{!}{%
\begin{tabular}{@{}p{0.35\textwidth}p{0.7\textwidth}@{}}
\toprule
\textbf{Process Parameter}                   & \textbf{Value}                                                                                      \\ \midrule
Axial depth of cut ($a_p$)                   & $12\;mm$                                \\
Radial depth of cut ($a_e$)                  & $1.2\;mm$                               \\
Cutting speed ($v_c$)                        & $435\;m/min$                            \\
Feed rate ($v_f$)                            &  $5540\;mm/min$                         \\
Feed per tooth ($f_z$)                       & $0.12\;mm$                              \\
Spindle speed  ($n$)                         & $11540\;rpm$                           \\
\bottomrule
\end{tabular}%
}
\caption{Process parameters from the measured acceleration data for the models}
\label{table:process_param}
\end{table*}



\twocolumn
\bibliographystyle{elsarticle-num} 
\bibliography{root}



\end{document}